\documentclass[preprint,5p,twocolumn]{elsarticle}

\usepackage{amssymb}
\usepackage{amsmath}
\DeclareMathOperator*{\argmax}{argmax}
\DeclareMathOperator*{\argmin}{argmin}
\providecommand{\norm}[1]{\lVert#1\rVert}

\usepackage{graphicx}
\graphicspath{{images/}}

\usepackage{subcaption}

\usepackage[]{algorithm2e}

\usepackage{hyperref}

\journal{Neurocomputing}

\begin{document}

% \linenumbers

\newcommand{\myfigure}[3] {
    \begin{figure}
        \centering
        \includegraphics[width=1.0\columnwidth]{#1}
        \caption{#2}
        \label{#3}
    \end{figure}
}

\newcommand{\vect}[1]{\boldsymbol{#1}}

\begin{frontmatter}


\title{Super-k: A Piecewise Linear Classifier Based on Voronoi Tessellations}

\author[1,3]{Rahman Salim Zengin
\corref{cor1}
\fnref{fn1}}
\ead{rszengin@itu.edu.tr}

\author[2,3]{Volkan Sezer
\fnref{fn2}}
\ead{sezerv@itu.edu.tr}

\cortext[cor1]{Corresponding author}

\fntext[fn1]{\href{https://orcid.org/0000-0002-3104-4677}{ORCID: https://orcid.org/0000-0002-3104-4677}}

\fntext[fn2]{\href{https://orcid.org/0000-0001-9658-2153}{ORCID: https://orcid.org/0000-0001-9658-2153}}

\address[1]{Department of Mechatronics Engineering, Istanbul Technical University, Istanbul, Turkey}
\address[2]{Department of Control and Automation Engineering, Istanbul Technical University, Istanbul, Turkey}
\address[3]{Autonomous Mobility Group, Electrical and Electronics Engineering Faculty, Istanbul Technical University, Istanbul, Turkey}

\begin{abstract}
    Voronoi tessellations are used to partition the Euclidean space into
    polyhedral regions, which are called Voronoi cells. Labeling the Voronoi
    cells with the class information, we can map any classification problem into
    a Voronoi tessellation. In this way, the classification problem changes into
    a query of just finding the enclosing Voronoi cell. In order to accomplish
    this task, we have developed a new algorithm which generates a labeled
    Voronoi tessellation that partitions the training data into polyhedral
    regions and obtains interclass boundaries as an indirect result. It is
    called Supervised k-Voxels or in short Super-k. We are introducing Super-k
    as a foundational new algorithm and opening the possibility of a new family
    of algorithms. In this paper, it is shown via comparisons on certain
    datasets that the Super-k algorithm has the potential of providing
    comparable performance of the well-known SVM family of algorithms with less
    complexity.
\end{abstract}


\end{frontmatter}

\section{Introduction}
\label{sec:introduction}

Piecewise Linear (PWL) classifiers are primitive and foundational types of
classifiers. When limited computational resources are considered, they become
appropriate tools for classification. To the best of our knowledge, almost all
piecewise linear classifier literature focuses on hyperplanes, margins, and
boundaries. It is very clear that these are critical concepts for the pattern
classification. A detailed literature review about PWL classifiers can be found
in\cite{leng_soft-margin_2020}.

Support Vector Machine (SVM)\cite{cortes_support-vector_1995} was the most used
classification algorithm before the deep learning
revolution\cite{carletti_age_2020}. SVM had been constructed based on strong
theoretical foundations. SVM obtains a separating hyperplane via choosing some
of the training data instances, which are maximizing the margin, as support
vectors. Although it is linear in the fundamental sense, with the help of kernel
trick, the classification problem is translated into a higher order feature
space, and nonlinear decision boundaries can be obtained for more complex
classification problems. When the kernel trick is utilized, SVM still
does its job as a linear classifier in that higher order feature space.

K-nearest neighbors (KNN)\cite{duda_pattern_2000} is a nonparametric method of
classification. Instead of obtaining a parametric model of the data, the data
itself is used as the model. When a new sample arrives for classification,
similar instances of the training data are found based on a distance metric,
mostly Euclidean distance. K-nearest training data instances constitutes a local
distribution, around the prediction point. Using the labels of the k data
instances, prediction can be done in many ways. Simple Bayesian statistics can
be applied, as the labels can be weighted equally, or weighted according to the
distances. Even a kernel function can be applied locally to determine the
predicted label. A famous quote summarizes KNN in a nice way: "When I see a bird
that walks like a duck and swims like a duck and quacks like a duck, I call that
bird a duck."\cite{riley_james_2017}

In this paper, a new machine learning algorithm, \textbf{Supervised k-Voxels} or
in short \textbf{Super-k}, is introduced. The Super-k algorithm comprises
several steps for processing the training data, and obtaining a classifier.
Voxelization\cite{hinks_point_2013} is applied to classes of data separately, in
order to obtain an initial Voronoi
tessellation\cite{okabeSpatialTessellationsConcepts2000}. A simple variant of EM
algorithm\cite{neal_view_1998} is applied to distribute initial generator points
over the data more uniformly. These steps result in separate Voronoi
tessellations for different classes. Afterwards, the Super-k algorithm merges
all separate class tessellations and labels the generator points, using the
covered data via plurality voting. In the final phase, it applies a simple
correction scheme to reduce False Positive (FP) classifications.

Due to the polyhedral shape of Voronoi cells, the Super-k algorithm is a PWL
classification algorithm. Generator points, which are labeled using the class
information, determine interclass boundaries. There is no direct consideration
of hyperplanes or boundaries in the proposed approach. On the other hand, these
hyperplanes and boundaries are indirect results of Voronoi-based space
partitioning.

As an enhancement, instead of Euclidean distance used in the definition of
Voronoi tessellations, derived Super-k likelihood is used as a similarity
metric. Maximization of the Super-k likelihood is equal to the minimization of
Euclidean distance (\ref{eq:superk_euclidean_equivalence}). Super-k likelihood
has a computational performance advantage on modern hardware due to the matrix
multiplication optimizations implemented in recent hardware platforms. It is
applicable to any algorithm that relies on Euclidean distance for one-to-many
comparisons.

The Super-k algorithm uses only simple arithmetic operations. Therefore, it can
be implemented using only integer arithmetic for both training and inference.
This enables the development of low-cost embedded platforms, not only for inference,
but also for training.

It is possible to split a pretrained Super-k model into multiple classifiers.
Similarly, separately trained Super-k models can be merged into a single model
to obtain higher precision, or higher number of classes. On-line or hybrid
learning scenarios can be implemented with this flexibility of the Super-k.

All alternate scenarios are left open and only the Super-k algorithm is
explained in this paper. Following a short background information (Section
\ref{sec:background}), the Super-k algorithm is explained (Section
\ref{sec:the_superk_algorithm}). Afterwards, experimental results,
classifications on synthetic data as well as comparisons against SVMs and KNN,
are given (Section \ref{sec:experimental}). Lastly, some ideas about possible
improvements and applications of the Super-k algorithm are shared (Section
\ref{sec:conclusion}).

\section{Background}
\label{sec:background}

In this section, notations and some definitions used throughout this paper are
provided. 

\subsection{Notation}
\label{sub:notation}

Boldface denotes a vector and superscript $^T$ denotes transpose, such as
$\vect{x} = (x_1, \ldots, x_m)^T$.

The floor function is shown as $\left\lfloor \cdot \right\rfloor$, and $\left\lceil
\cdot \right\rceil$ is the ceil function. Rounding to the nearest integer is shown
as $\left\lfloor \cdot \right\rceil$.

As piecewise linear (PWL) classification is done on multivariate data, Voronoi
Tessellations\cite{okabeSpatialTessellationsConcepts2000} covering such data
are defined in $m$ dimensional Euclidean space, $\mathbb{R}^m$.

A Voronoi polyhedron is a convex region defined by an inner generator point
and some outer generator points, as shown in (\ref{eq:voronoi_polyhedon}).

\begin{equation}
    \begin{split}
        V(\vect{p_i}) = \{ & \vect{x} |\ \|\vect{x} - \vect{p_i}\| \le 
                                  \|\vect{x} - \vect{p_j}\|\\
                    & \text{for}\ j \neq i,\ i,j \in I_n = \{1,\ldots,n\} \}
    \end{split}
    \label{eq:voronoi_polyhedon}
\end{equation}
where $V(\vect{p_i})$ denotes voronoi polyhedron with respect to the generator
point $\vect{p_i}$.

The set of Voronoi polyhedra which creates the Voronoi tessellation is defined as
$\mathcal{V} = \{ V(\vect{p_1}), \ldots, V(\vect{p_n}) \}$.

A generator point $\vect{p_i}$ belongs to the set of generator points $P$ of the
Voronoi tessellation, hence, the set of generator points is 
$P = \{\vect{p_1}, \vect{p_2}, \ldots, \vect{p_n})$.

Voronoi facets are sets of equidistant points between generator points as
given in \ref{eq:facet},

\begin{equation}
    e_{ij} = \{ \vect{x} |\ \|\vect{x} - \vect{p_i}\| = \|\vect{x} - \vect{p_j}\| \}
    \label{eq:facet}
\end{equation}
where $j \neq i$.

The whole set of facets of a Voronoi polyhedron is called a boundary and 
it is denoted related to the inner generator point of the region encircled as 
$\partial V(\vect{p_i})$.

Labeling the generator points set $P$ results in a labeled tessellation. Every
Voronoi cell has a class designator $\xi_i$, which is defined with respect to
$\vect{p_i}$. Set of all class designators of the generator points is defined as
$\Xi = \{ \xi_1, \ldots, \xi_n \}$.

Superscript $^{(t)}$ denotes the value of a parameter for the iteration step of $t$.

In the classification plots, misclassified data instances are marked with red
dots.

\subsection{The Super-k Likelihood}
\label{sub:super-k_likelihood}

If the logarithm of the multivariate normal distribution is simplified
(\ref{apx:likelihood_derivation}) for $\Sigma = I$ and equal priors are
considered, for the maximum likelihood approximation, the derived likelihood
function is given in (\ref{eq:likelihood}).

\begin{equation}
    g_i(\vect{x}) = \vect{x}^T \vect{p_i} - \frac{1}{2} \vect{p_i}^T \vect{p_i}
    \label{eq:likelihood}
\end{equation}

In order to understand the usage of (\ref{eq:likelihood}), it can be separated
into two parts as given in (\ref{eq:likelihood_separated}).

\begin{equation}
    g_i(\vect{x}) = \underbrace{\vect{x}^T \vect{p_i}}_{\text{changing}}
    - \underbrace{\frac{1}{2} \vect{p_i}^T \vect{p_i}}_{\text{constant}}
    \label{eq:likelihood_separated}
\end{equation}

After the training of a Super-k classifier, the generator points become constant.
Then, the second part of (\ref{eq:likelihood_separated}) is composed of just
constant numbers, and computing them at once for a trained model is enough.
Hence, the computation of the Super-k likelihood is a reduction to the inner product of
two vectors, plus a constant. The computation of the Super-k likelihood can be done
with a single \textit{Basic Linear Algebra Subprograms}
(BLAS)\cite{blackford_blas_1999} Level 1 function call. Almost every modern
hardware platform has vector computation capabilities which makes them very
suitable for the acceleration.

\section{The Super-k Algorithm}
\label{sec:the_superk_algorithm}

In order to convert the training data to a Voronoi Tessellation-based
representation, the spaces, which are covering the whole range of every class of
the data, are divided into voxels separately per every class. Instances of the
training data are partitioned into voxelized subsamples through this process.
Some of the voxels might be empty, some others might include varying numbers of
instances. For every nonempty voxel, the mean of the included instances is
calculated. These voxel means create the initial tessellation of the class and
represent the generator points of the class.

Tessellating the data through a uniform voxelization process might result in a
nonideal tessellation, and possibly creates degenerate\footnote{For the degenerate
cases see page 46 of \cite{okabeSpatialTessellationsConcepts2000}} Voronoi
vertices. The generator points and class data are passed through a simple
Expectation-Maximization process to evenly distribute the generator points over
the class instances. In this way, possible degenerate Voronoi vertices are
removed, and the generator points move closer to the centroids of the Voronoi
cells.

All separately created Voronoi tessellations are merged into a single
tessellation via combining the generator points of every class. When all the
generators are combined within some overlapping regions between classes, some
of the generator points might become minorities, compared to the enclosed data
instances. In order to make sure that all of the generator points represent the
assigned instances correctly, labeling via plurality voting is performed.

Finally, with the goal of reducing FP classifications, a correction is applied to
the generator points via excluding FP instances from the relevant Voronoi cells.
This correction reduces the classification error without causing overfitting of the
training data.

\begin{figure}[h]
    \centering
    \includegraphics[width=0.6\linewidth]{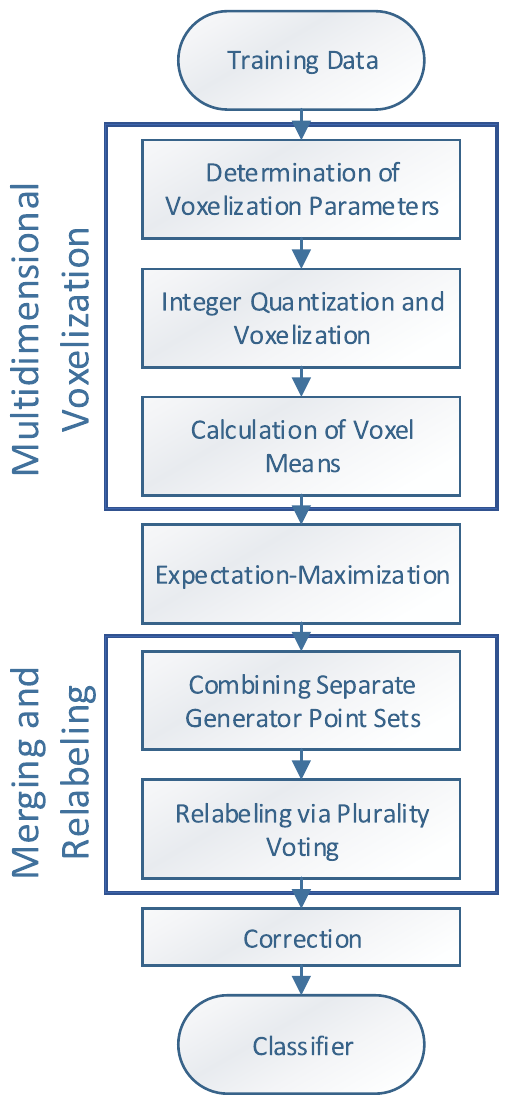}
    \caption{The Super-k algorithm processing steps}
    \label{fig:processing_steps}
\end{figure}

The main steps of the Super-k algorithm is shown in (Figure
\ref{fig:processing_steps})). Each of these steps are explained in detail, in
the following sections.

\subsection{Multidimensional Voxelization}
\label{sub:multi-d_voxelization}

Voxelization of multidimensional data is not as simple as doing so with 3-D
data. Dividing the space, covered with the whole range of data using all
dimensions might create impractically a high number of voxels. For
$\mathbb{R}^m$, if every dimension is divided into $b$ ranges, the number of
total voxels would be $b^m$. For a sample size of $N$, there can be at most $N$
nonempty voxels. However, having just one data instance in every nonempty voxel
merely gives back the original data, after the voxelization process. For a
meaningful voxelization, some of the nonempty voxels must have multiple data
instances. Therefore, the number of nonempty voxels must be less than the sample
size $N$. Thus, when a uniformly distributed data is considered, the
practicability condition of voxelization can be defined as $b^m < N$ or via
taking the logarithm of both sides with respect to $b$, it becomes $m <
\log_{b}{N}$.

In order to accomplish the task of voxelization of multidimensional data, an
empirical method is developed. $k$ is an indirect parameter which specifies the
minimum number of required voxels. It is not a precision parameter and its
effect depends on the data. Multiple values of $k$ might give the same result.
In order to understand the reasoning behind our assumptions, a numerical example
will be provided. Afterwards, relevant equations will be given.

Let us assume that for a given dataset, the sample size of the data is $N=1000$,
the dimensionality of the data is $m=100$, and the requested number of voxels is
$k=50$. We need to calculate $b$ in order to find how many divisions are
required for each axes. The value of $b$ can only be an integer. Thus, let us
define another real number parameter $c$, the scale of divisions, and relate it
to $b$ as $b = \left\lceil c \right\rceil$. From the definition of $k$, we can
write $c^m = k$ and obtain $c$ using $k$ as $c = \sqrt[m]{k}$. Using the
numerical values provided in the example, the value of $c$ can be found as $c =
\sqrt[100]{50} = 1.039896$. It is not possible to divide the dimensions into
fractional ranges, hence, this number is not usable. If we round this number to
the next higher integer value, $b$ is found as $b = \left\lceil c \right\rceil =
2$. This value of $b$ is usable for the voxelization. However, now, the number
of voxels becomes $b^m = 2^{100} = 1267650600228229401496703205376$. This number
is higher than all the memories in the world.

If we use not all, but a small subset of the dimensions for the voxelization, it
will be possible to obtain a more reasonable number of voxels. Let us designate
the number of dimensions to use as $m_v$, which we name as the number of variant
features. Furthermore, let us introduce another parameter, $a = \left\lfloor
c\right\rfloor $, which we use for the number of divisions of the unselected
dimensions. Relation of $b$ with $c$ is $b = \left\lceil c \right\rceil$. With
these definitions, our aim is to approximate $c$ with the integers $a$ and $b$,
such that, $a^{(m - m_v)} b^{m_v} \approx c^m$. Then, $m_v$ can be given as,

\begin{equation}
    m_v = \left\lfloor {m \frac{\log{\left(\frac{c}{a}\right)}}{\log{\left(\frac{b}{a}\right)}}}\right\rceil
    \label{eq:calculation_mv}
\end{equation}
Derivation of (\ref{eq:calculation_mv}) can be found in (\ref{apx:derive_mv}).

For our numerical example, when we do the calculations using
(\ref{eq:calculation_mv}), the result becomes $m_v = \left\lfloor 5.643856
\right\rceil = 6 \label{eq:numeric_mv}$.

The selection of the variant features is another critical issue. Although the
variance shows spread of data, it cannot measure how uniformly the data spreads
over the existing values. Under these considerations, the number of unique
values for every feature is used as a measure for the selection of the variant
features. The higher the number of the unique values is, the better it makes the
voxelization.

\begin{figure}
    \centering
    \begin{subfigure}[t]{0.99\linewidth}
        \centering
        \includegraphics[width=0.7\textwidth]{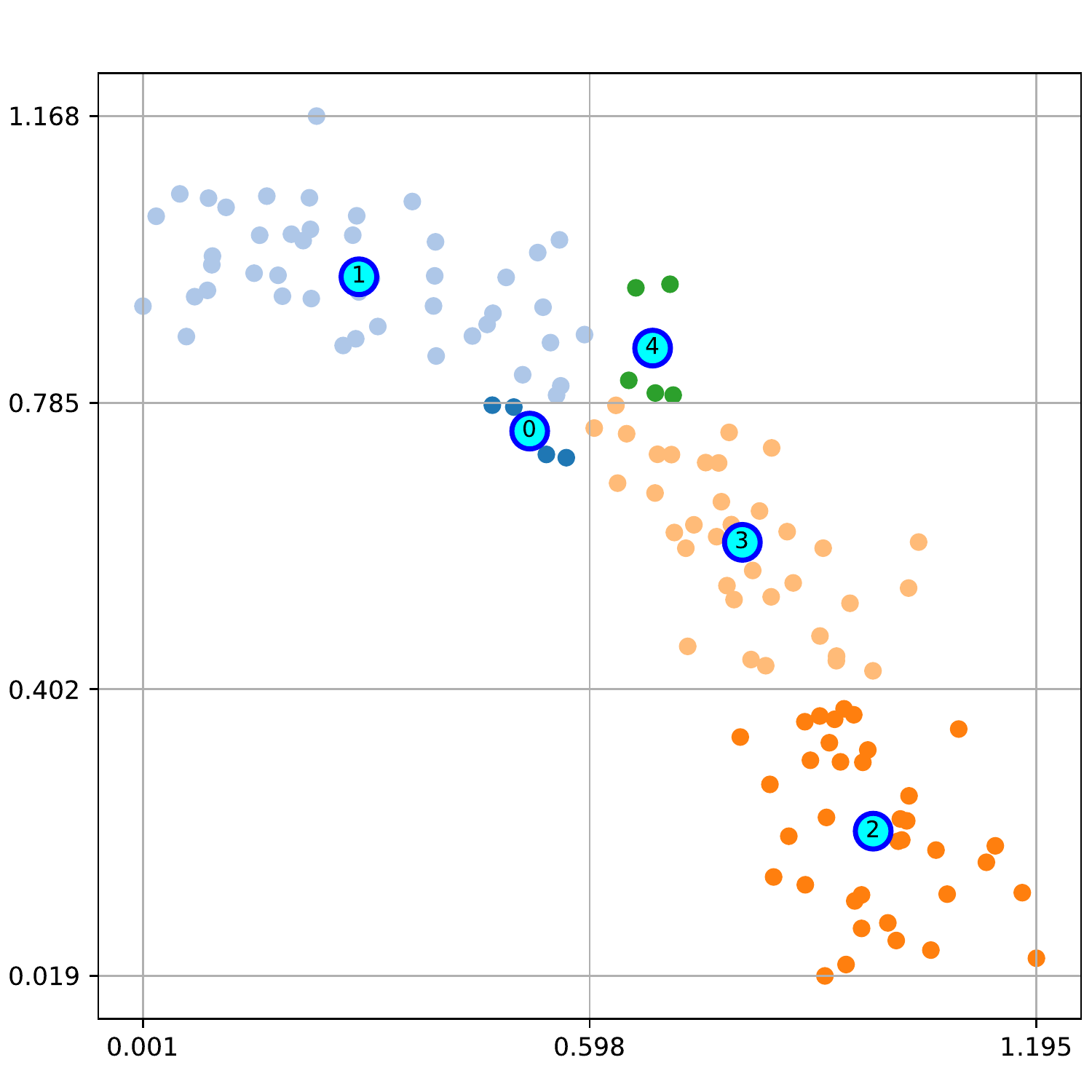}
        \caption{Class 0}
        \label{fig:voxelization:class_0}
    \end{subfigure}
    \begin{subfigure}[t]{0.99\linewidth}
        \centering
        \includegraphics[width=0.7\textwidth]{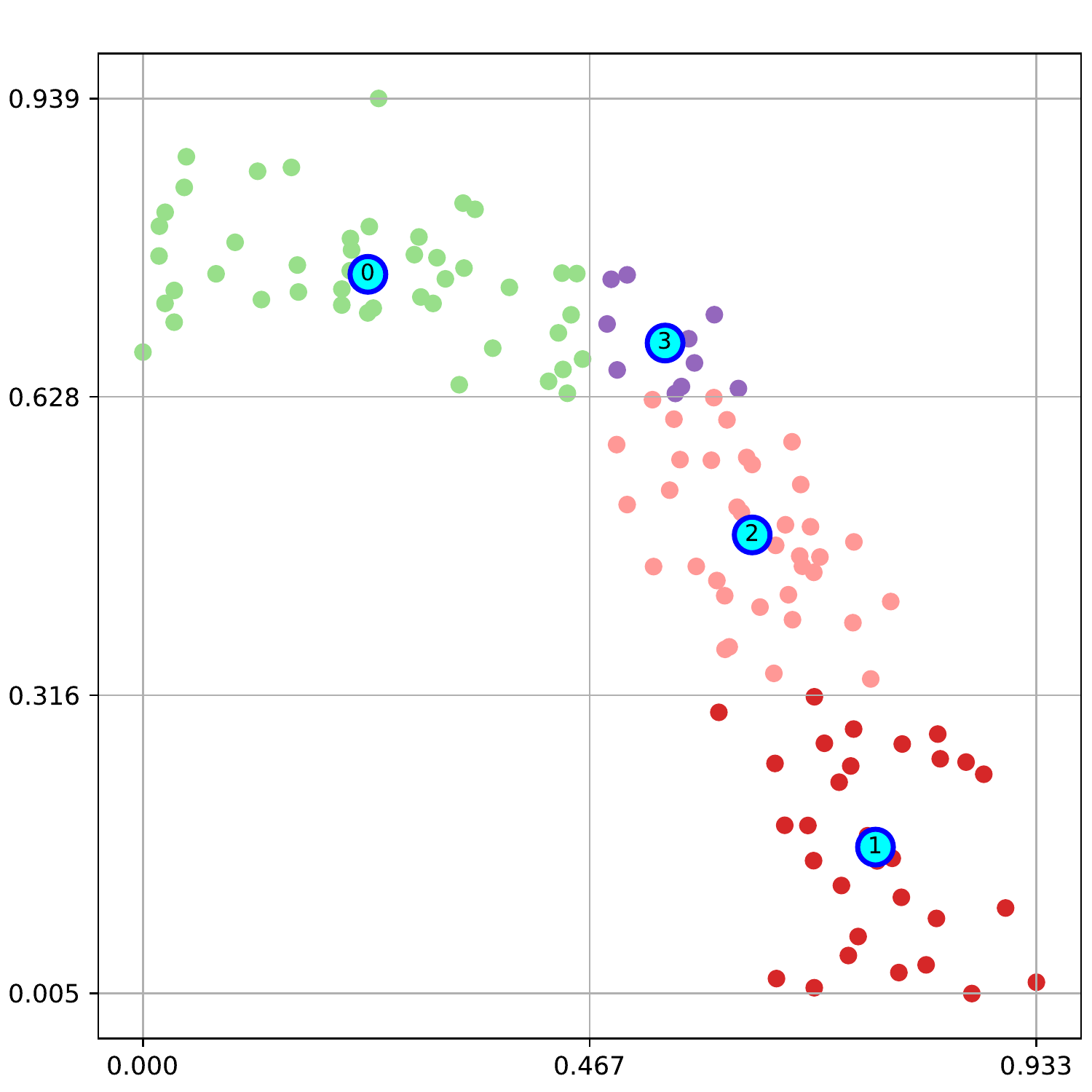}
        \caption{Class 1}
        \label{fig:voxelization:class_1}
    \end{subfigure}
    \caption{Voxelization grids, voxel members, and voxel means}
    \label{fig:voxelization}
\end{figure}

\begin{figure}
    \centering
    \begin{subfigure}[t]{0.82\linewidth}
        \centering
        \includegraphics[width=\textwidth]{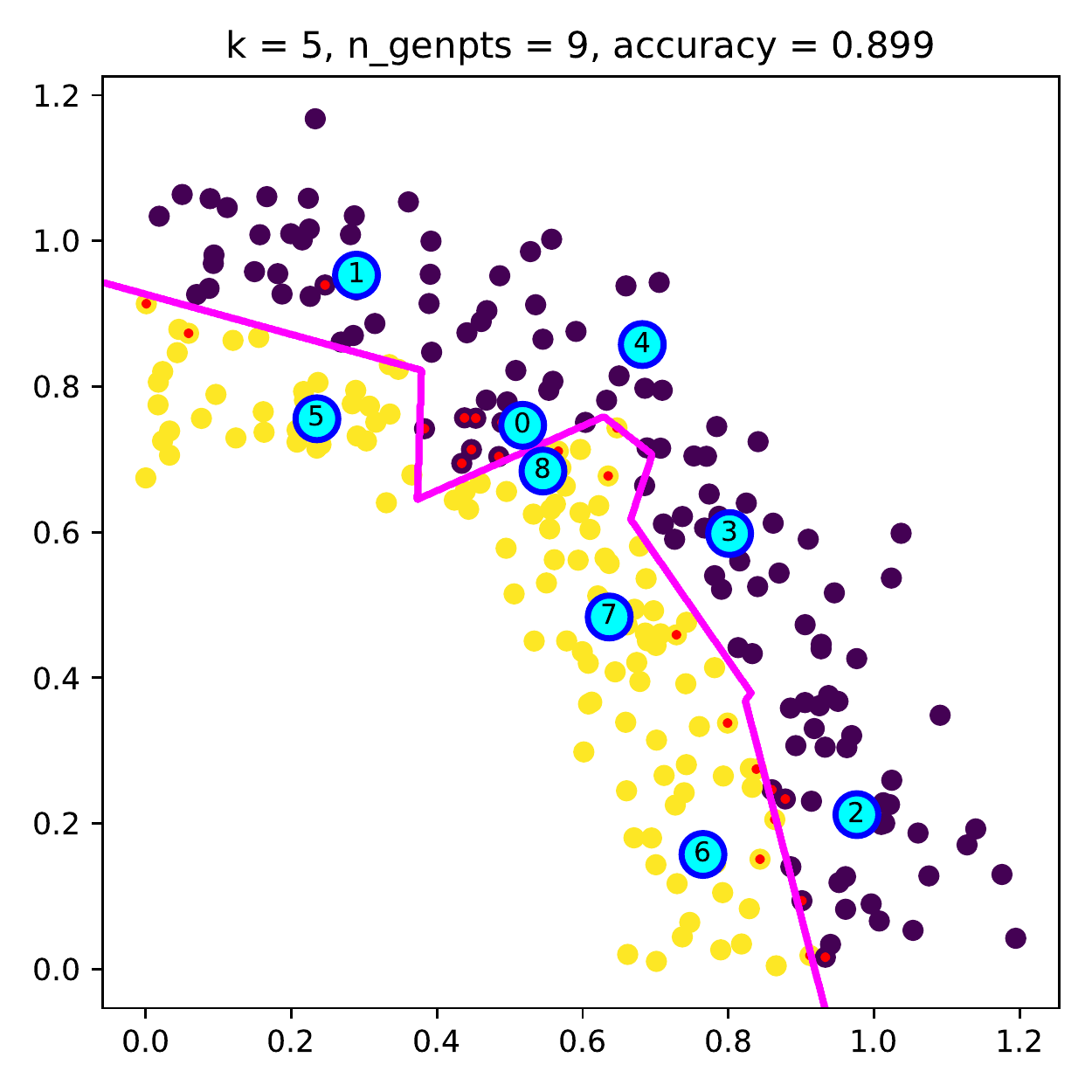}
        \caption{Voxelization}
        \label{fig:train:voxel}
    \end{subfigure}
    \begin{subfigure}[t]{0.82\linewidth}
        \centering
        \includegraphics[width=\textwidth]{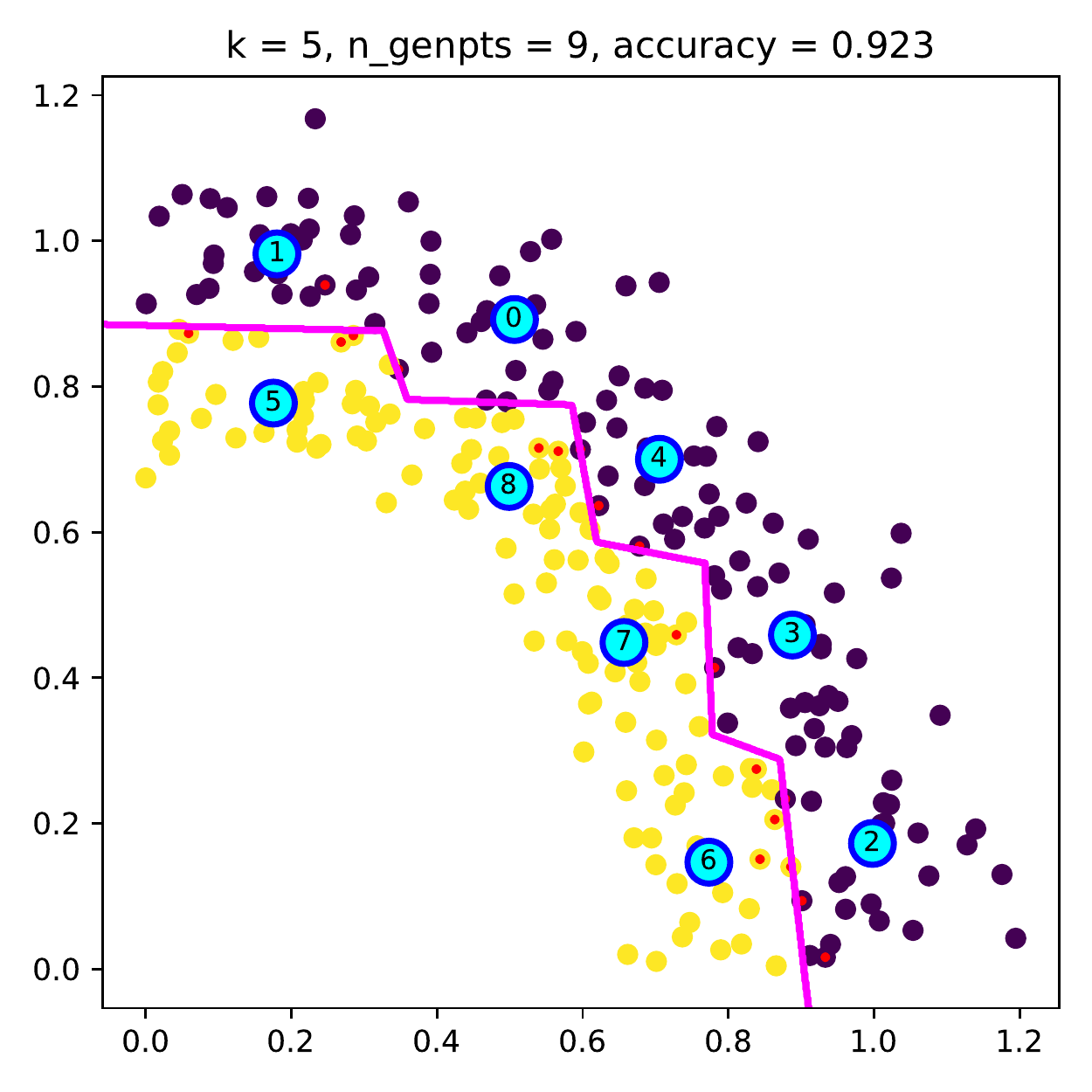}
        \caption{EM}
        \label{fig:train:em}
    \end{subfigure}
    \begin{subfigure}[t]{0.82\linewidth}
        \centering
        \includegraphics[width=\textwidth]{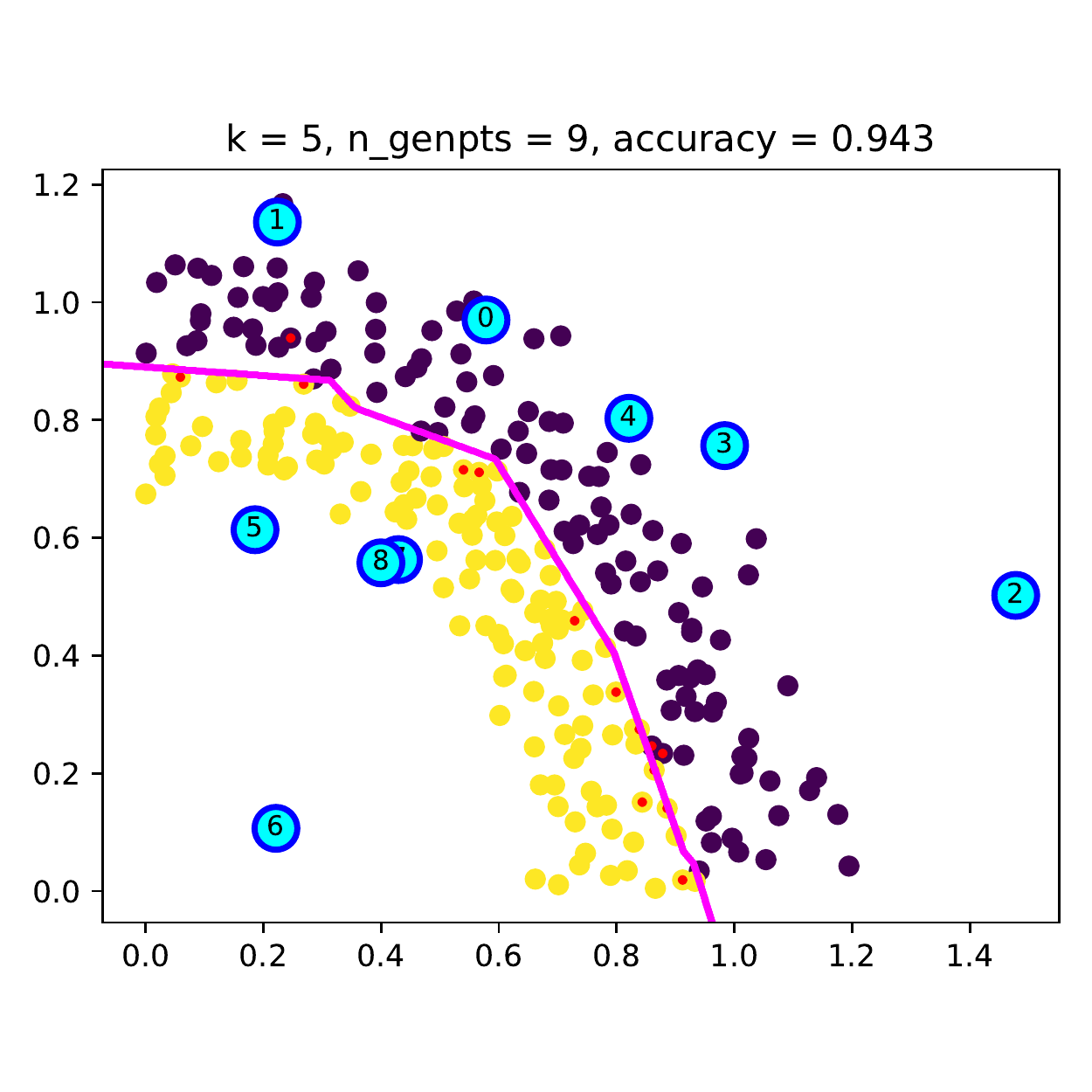}
        \caption{Correction}
        \label{fig:train:correct}
    \end{subfigure}
    \caption{Training steps of the Super-k algorithm}
    \label{fig:train}
\end{figure}

The final step of the voxelization is the determination of the same voxel
instances via quantization and voxel means calculation. Min to max ranges of the
data is divided into the number of steps, separately for every data feature.
Using this resolution parameter, the whole data is converted to integer indices,
which designate the voxel indices of the data instances. Then, the means of all
nonempty voxels are calculated using the instances inside. The related
pseudocode is given in (Algorithm \ref{alg:voxelization}). The voxelization
process is also illustrated in (Figure \ref{fig:voxelization}), where
voxelization of two classes is shown separately. For $k=5$, the algorithm
determines $3 \times 2$ partitioning of the whole range for both classes. For
class 0 (Figure \ref{fig:voxelization:class_0}), there are 5 nonempty voxels.
For class 1 (Figure \ref{fig:voxelization:class_0}), the number of nonempty
voxels is 4. Data instances in separate voxels are shown in different colors.
The voxel means are also shown as cyan circles with voxel indices inside. These
two separate results can be seen combined in (Figure \ref{fig:train:voxel}).

\begin{algorithm}
    \DontPrintSemicolon
    \SetKwProg{Fn}{Function}{}{}
    \SetKwFunction{Vox}{Voxelize}
    \Fn(){\Vox{k}}{}{
        \KwData{\\
            $X \longleftarrow \begin{bmatrix}
                \vect{x_1} & \cdots & \vect{x_N}
            \end{bmatrix}$ \tcc{Instances}
        }
        \KwResult{$Means$ \tcc*[h]{Voxel means}}
        \Begin{
            $c = \sqrt[m]{k};\ 
            a = \left\lfloor c \right\rfloor\;\ 
            b = \left\lceil c \right\rceil$\;
            $m_v = \left\lfloor m \frac{\log{\left(\frac{c}{a}\right)}}{\log{\left(\frac{b}{a}\right)}}\right\rceil$\;
            \For(\tcc*[h]{every dimension}){$i = {1, \ldots, m}$}{
                $Counts[i] \longleftarrow$ CountUniqueValues
            }
            \tcc{Get indices of $m_v$ maximum}
            $DimIndices$ $\longleftarrow$ ArgK-Max($Counts$, $m_v$)\;
            $X_v \longleftarrow$ GetDims($X, DimIndices$)\;
            \tcc{Quantize over selected dimensions}
            $Indices \longleftarrow$ Quantize($X_v$)\;
            \tcc{Map voxel indices back to the data and calculate voxel means}
            $Means \longleftarrow$ CalculateMeans($X, Indices$)\;
        }
    }
    \caption{Multidimensional Voxelization}
    \label{alg:voxelization}
\end{algorithm}

\subsection{Cycling Through EM}
\label{sub:em_cycling}

A very simple variant of the Expectation-Maximization (EM) algorithm is used to
distribute generator points over the class instances, more uniformly.

The Super-k likelihood (\ref{eq:likelihood}) is used for the maximization step. The
expectation step is the calculation of the means of the instances assigned to the
same generator points. The effect of applying EM can be seen if (Figure
\ref{fig:train:voxel}) is compared with (Figure \ref{fig:train:em}). The
generator point $0$ moves up, and the generator points $4$ and $3$ move downright.
As a result, the accuracy of the classification is improved from 89.9\% to
92.3\% at the end of this step, as shown in (Figure \ref{fig:train}). Pseudocode
of the EM variant is given in (Algorithm \ref{alg:em_cycling}).

\begin{algorithm}
    \DontPrintSemicolon
    \SetKwProg{Fn}{Function}{}{}
    \SetKwFunction{Em}{ApplyEM}
    \Fn(){\Em{$n_{cycles}$}}{}{
        \KwData{\\
            $X \longleftarrow \begin{bmatrix}
                \vect{x_1} & \cdots & \vect{x_N}
            \end{bmatrix}$ \tcc{Instances}
            $P \longleftarrow \begin{bmatrix}
                \vect{p_1} & \cdots & \vect{p_n}
            \end{bmatrix}$ \tcc{Generator Points}
        }
        \KwResult{$P$: \tcc*[h]{Generator Points}}
        \Begin{
            \For(){$i = {1, \ldots, n_{cycles}}$}{
                \tcc{Assign data instances to the generator points}
                $Assignments$ = Argmax(Likelihood($X, P$))\;
                \tcc{Recalculate generator points from the assigned instances}
                \ForEach{$p_i$ in $P$}{
                    $p_i$ $\longleftarrow$ Mean($X[Assignments = i]$)\;
                }
            }
        }
    }
    \caption{Simple EM variant}
    \label{alg:em_cycling}
\end{algorithm}

\subsection{Merging and Relabeling}
\label{sub:merge_and_relabel}

In order to create a single Voronoi tessellation from separately created class
tessellations, all generator points should be merged and labeled according to the
pluralities of the data instances covered. The merging of two separate classes
can be understood via examination of (Figure \ref{fig:voxelization}) and (Figure
\ref{fig:train:voxel}). The difference of plurality voting from majority voting is
that plurality has the highest number of votes but might be less than half of all
votes, while majority takes more than half of all votes. When multiple classes are
considered, plurality voting is more suitable. Related pseudocode is given in
(Algorithm \ref{alg:merge_label}).

\begin{algorithm}
    \DontPrintSemicolon
    \SetKwProg{Fn}{Function}{}{}
    \SetKwFunction{MergeAndRelabel}{MergeAndRelabel}
    \Fn(){\MergeAndRelabel{$n_{classes}$}}{}{
        \KwData{\\
            $X \longleftarrow \begin{bmatrix}
                \vect{x_1} & \cdots & \vect{x_N}
            \end{bmatrix}$ \tcc{Instances}
            $y \longleftarrow \begin{bmatrix}
                y_1 & \cdots & y_N
            \end{bmatrix}$ \tcc{Labels}
            \tcc{$\mathcal{P}$: Set of Class Generators}
            $\mathcal{P} \longleftarrow \{P_1, \ldots, P_{n_{classes}}\}$\;
        }
        \KwResult{\\
            $P$: \tcc*[h]{Combined Generator Points}\;
            $\Xi$: \tcc*[h]{Generator Point Labels}
        }
        \Begin{
            $P$ $\longleftarrow$ Combine($\mathcal{P}$)\;
            \tcc{Assign data instances to the generator points}
            $Assn$ = Argmax(Likelihood($X, P$))\;
            \ForEach{$p_i$ in $P$}{
                $Counts$ $\longleftarrow$ CountUnique($y[Assn = i]$)\;
                $\xi_i$ $\longleftarrow$ Argmax(Counts)\;
            }

        }
    }
    \caption{Merging and Labeling of the combined tessellation}
    \label{alg:merge_label}
\end{algorithm}

\subsection{Correction}
\label{sub:correction}

Every Voronoi cell is defined by its generator point. Due to the labeling of the
generator points, some of the data instances in the relevant Voronoi cell, might
belong to another class. Therefore, these instances are the FPs of that Voronoi cell.
If these FP instances are close to the cell boundary, moving the generator point
away from the FP instances will leave them out of the Voronoi cell. In order
to apply this simple correction idea, the generator points are weighted with the number
of all assigned instances and the FP instances are subtracted from the generator
points. Repetitive application of this correction (\ref{eq:correction}) improves
classification accuracy without causing overfitting. An example result of
correction is shown in (Figure \ref{fig:train:correct}). If it is compared with
the uncorrected state (Figure \ref{fig:train:em}), change of the decision
boundary is apparent. The generator points move away from the boundary, and the
boundary becomes smoother. As a result, the accuracy is improved from 92.3\% to
94.3\% for the provided example.

\begin{equation}
    \vect{p_i}^{(t+1)} = \frac{\vect{p_i}^{(t)} * n_{all} - \sum \vect{x_{FP}} }{n_{all} - n_{FP}}
    \label{eq:correction}
\end{equation}

The pseudocode of the correction procedure is given in (Algorithm \ref{alg:corection}).

\begin{algorithm}
    \DontPrintSemicolon
    \SetKwProg{Fn}{Function}{}{}
    \SetKwFunction{Correction}{Correction}
    \Fn(){\Correction{$n_{cycles}$}}{}{
        \KwData{\\
            $X \longleftarrow \begin{bmatrix}
                \vect{x_1} & \cdots & \vect{x_N}
            \end{bmatrix}$ \tcc{Instances}
            $P \longleftarrow \begin{bmatrix}
                \vect{p_1} & \cdots & \vect{p_n}
            \end{bmatrix}$ \tcc{Generator Points}
        }
        \KwResult{$P_{best}$: \tcc*[h]{Generator Points}}
        \Begin{
            $best\_accuracy = 0$\;
            \For(){$i = {1, \ldots, n_{cycles}}$}{
                \ForEach{$p_i$ in $P$}{
                    $p_i = (p_i * n_{all} - \sum x_{FP}) / (n_{all} - n_{FP})$\;
                }
                \tcc{Keep best params}
                \If{$accuracy > best\_accuracy$}{
                    $best\_accuracy = accuracy$\;
                    $P_{best} = P$
                }
            }
        }
    }
    \caption{Correction}
    \label{alg:corection}
\end{algorithm}

\subsection{Classification with Super-k}
\label{sub:classification}

Classification using the trained Super-k model is the maximization of the Super-k
likelihood over $i$ (\ref{eq:classify_single}). This maximization is
straightforward to implement with a few lines of code. When
(\ref{eq:classify_single}) is applied, the class label of the $i$'th generator point
becomes the classification result. The advantage of the Super-k likelihood is
mentioned in (Section \ref{sub:super-k_likelihood}). Its relation to the
Euclidean distance is explained in (\ref{apx:likelihood_derivation}).

\begin{equation}
    \xi(x) = \argmax_{i}{\left( \vect{x}^T \vect{p_i} - \frac{1}{2} \vect{p_i}^T \vect{p_i} \right)}
    \label{eq:classify_single}
\end{equation}

\section{Experimental Results}
\label{sec:experimental}

The platform used for the experimental tests and comparisons is as follows: The
CPU used for the experimentation is Intel(R) Core(TM) i7-7700 running at 3.60GHz
frequency; the system has 32GB of DDR4 RAM running at 2133 MHz.

SVM and KNN are inherently PWL classifiers. SVM chooses some near the boundary
data instances. KNN creates Voronoi-like partitioning. Classification with
Super-k is similar to 1-NN classification. Due to such similarities, the Super-k
algorithm is compared against these algorithms.

Unoptimized reference implementation of the Super-k algorithm is written in
Python\cite{van_rossum_python_1995} using the related
libraries\cite{van_der_walt_numpy_2011,hunter_matplotlib_2007}.
Scikit-Learn\cite{pedregosa_scikit-learn_2011} is used for both synthetic data
generation and dataset retrieval. In addition, the SVM variants and KNN are used
from the same library\cite{pedregosa_scikit-learn_2011}. According to the
library documentation, for Linear-SVM, LibLinear\cite{fan_liblinear_2008} is
used in the background. Besides, LibSVM is used for the kernel based SVMs by the
library.

The reference implementation of the Super-k algorithm and the source code to
reproduce the results of this paper are
shared\cite{zengin_ituamgsuper-k_2021}.

\subsection{Tests with synthetic datasets}
\label{sub:synthetic_tests}

\begin{figure*}
    \centering
    \begin{subfigure}[b]{0.32\textwidth}
        \centering
        \includegraphics[width=\textwidth]{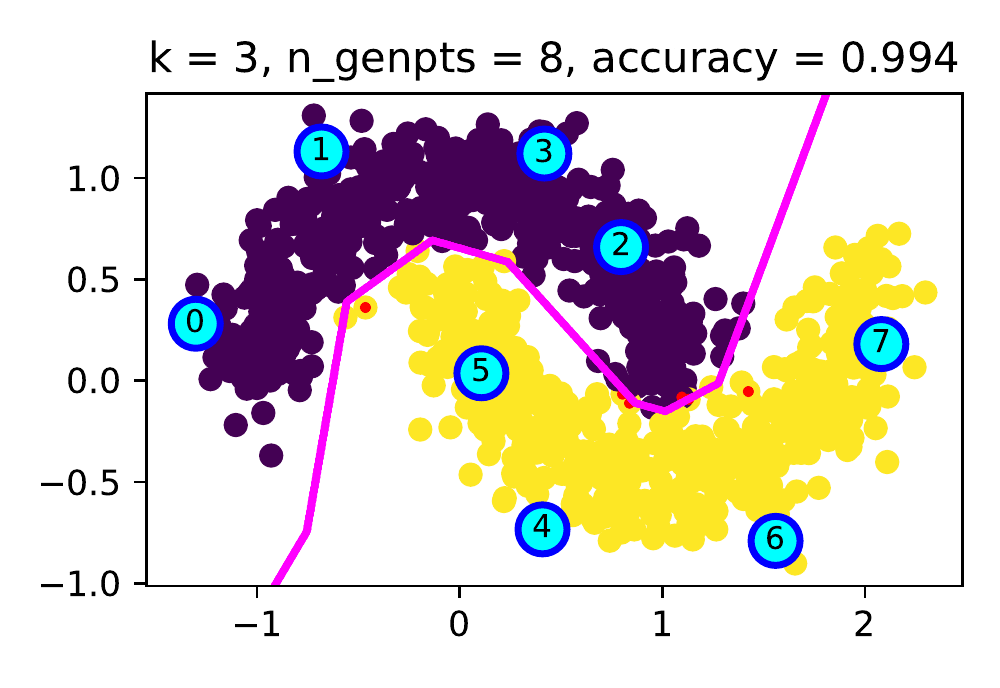}
        \caption{$k=3$}
        \label{fig:test_moons:1}
    \end{subfigure}
    \begin{subfigure}[b]{0.32\textwidth}
        \centering
        \includegraphics[width=\textwidth]{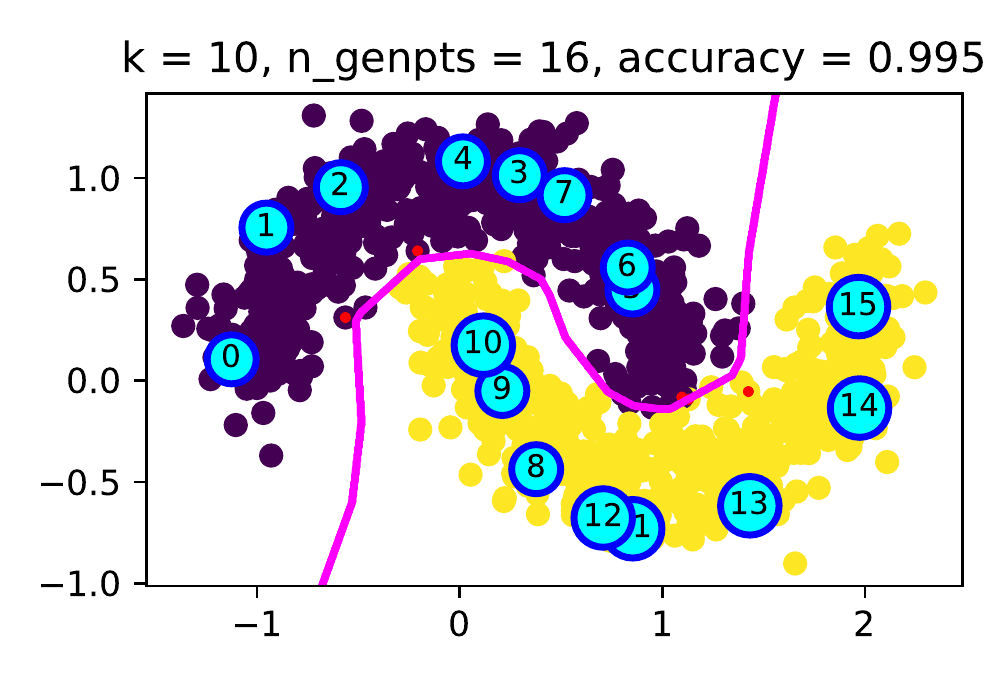}
        \caption{$k=10$}
        \label{fig:test_moons:2}
    \end{subfigure}
    \begin{subfigure}[b]{0.32\textwidth}
        \centering
        \includegraphics[width=\textwidth]{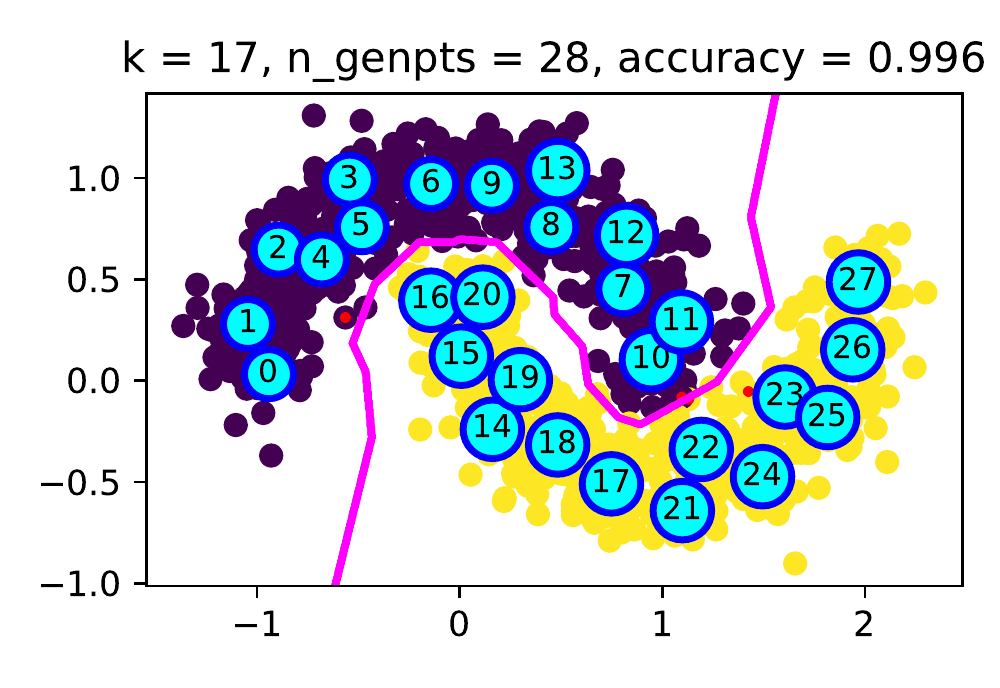}
        \caption{$k=17$}
        \label{fig:test_moons:3}
    \end{subfigure}
    \caption{Test results on synthetic moons dataset}
    \label{fig:test_moons}
\end{figure*}

\begin{figure*}
    \centering
    \begin{subfigure}[b]{0.32\textwidth}
        \centering
        \includegraphics[width=\textwidth]{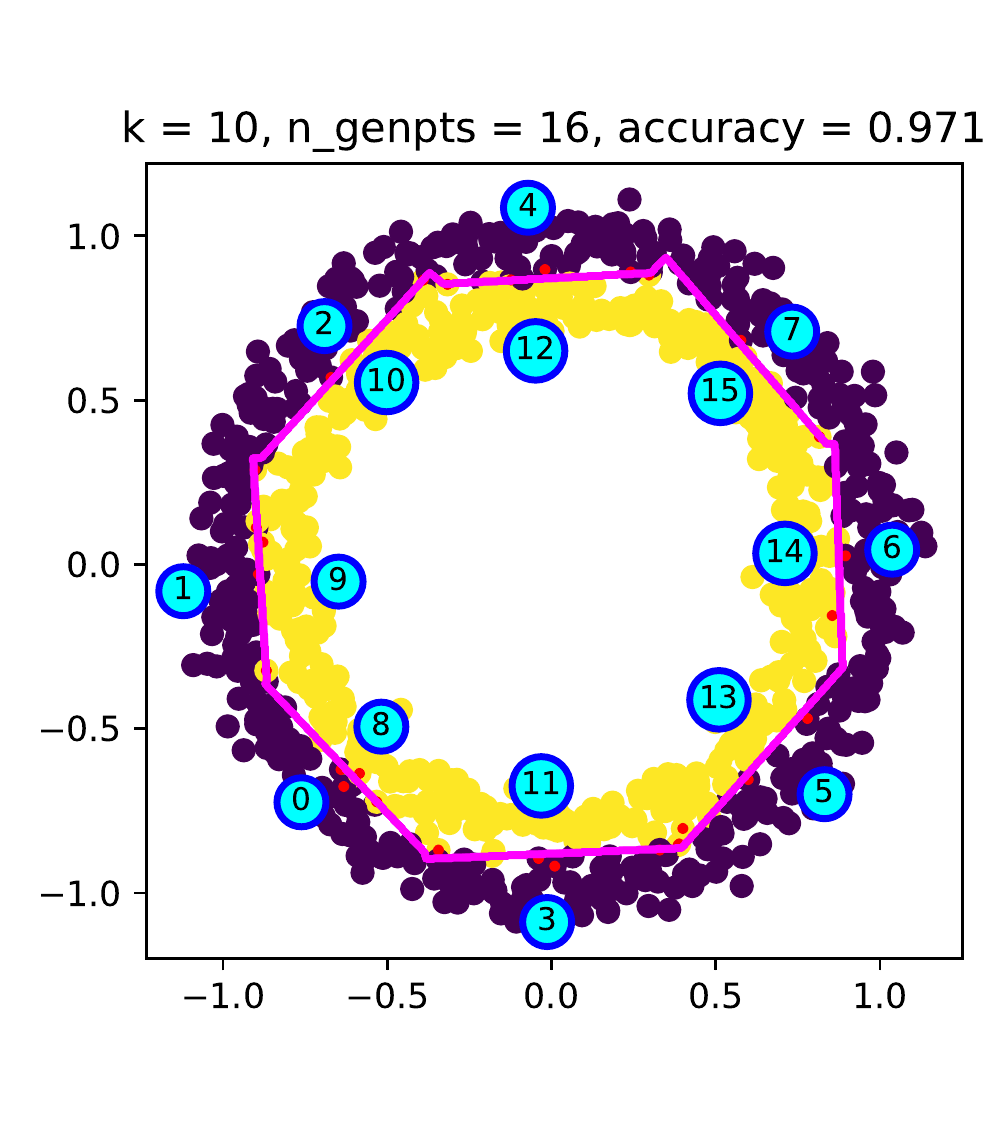}
        \caption{$k=10$}
        \label{fig:test_circles:1}
    \end{subfigure}
    \begin{subfigure}[b]{0.32\textwidth}
        \centering
        \includegraphics[width=\textwidth]{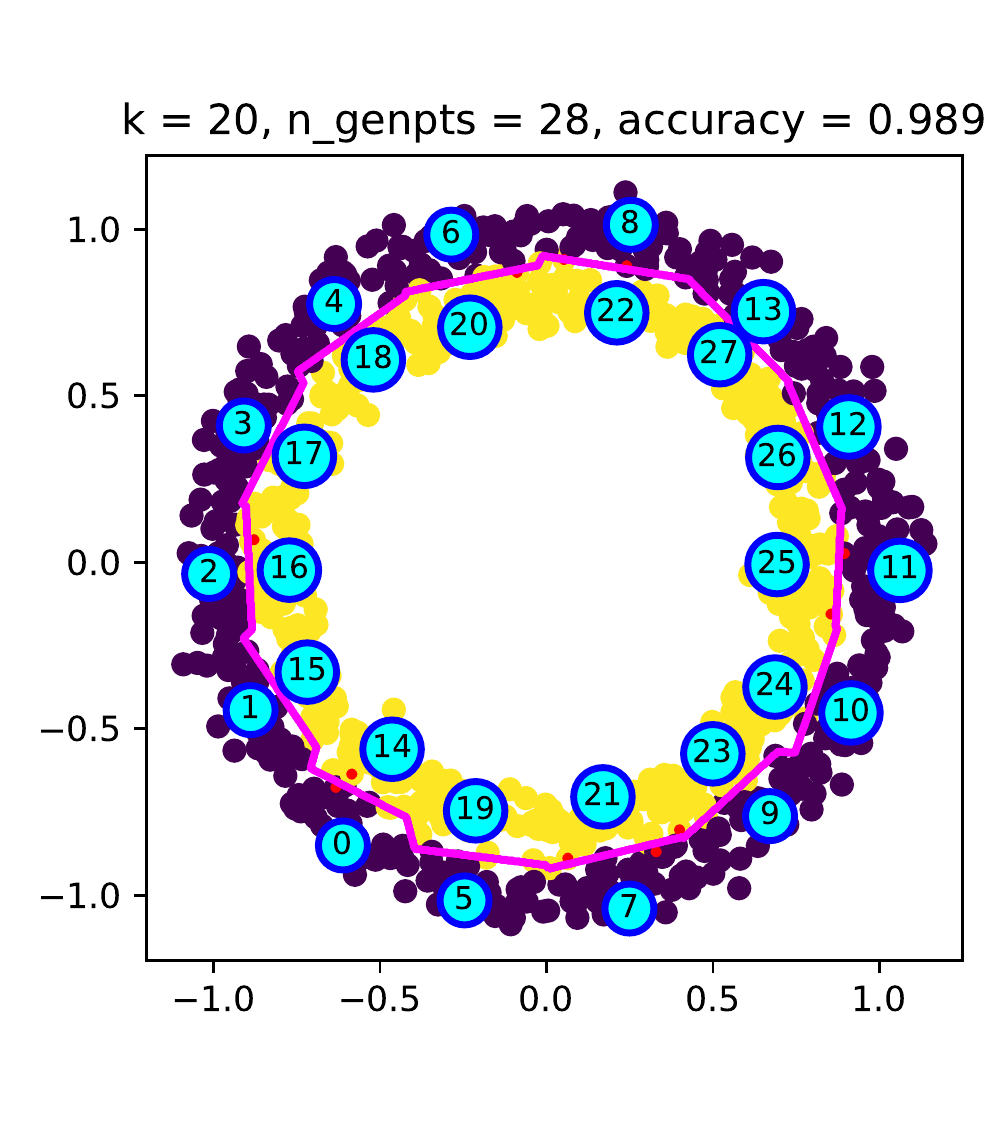}
        \caption{$k=20$}
        \label{fig:test_circles:2}
    \end{subfigure}
    \begin{subfigure}[b]{0.32\textwidth}
        \centering
        \includegraphics[width=\textwidth]{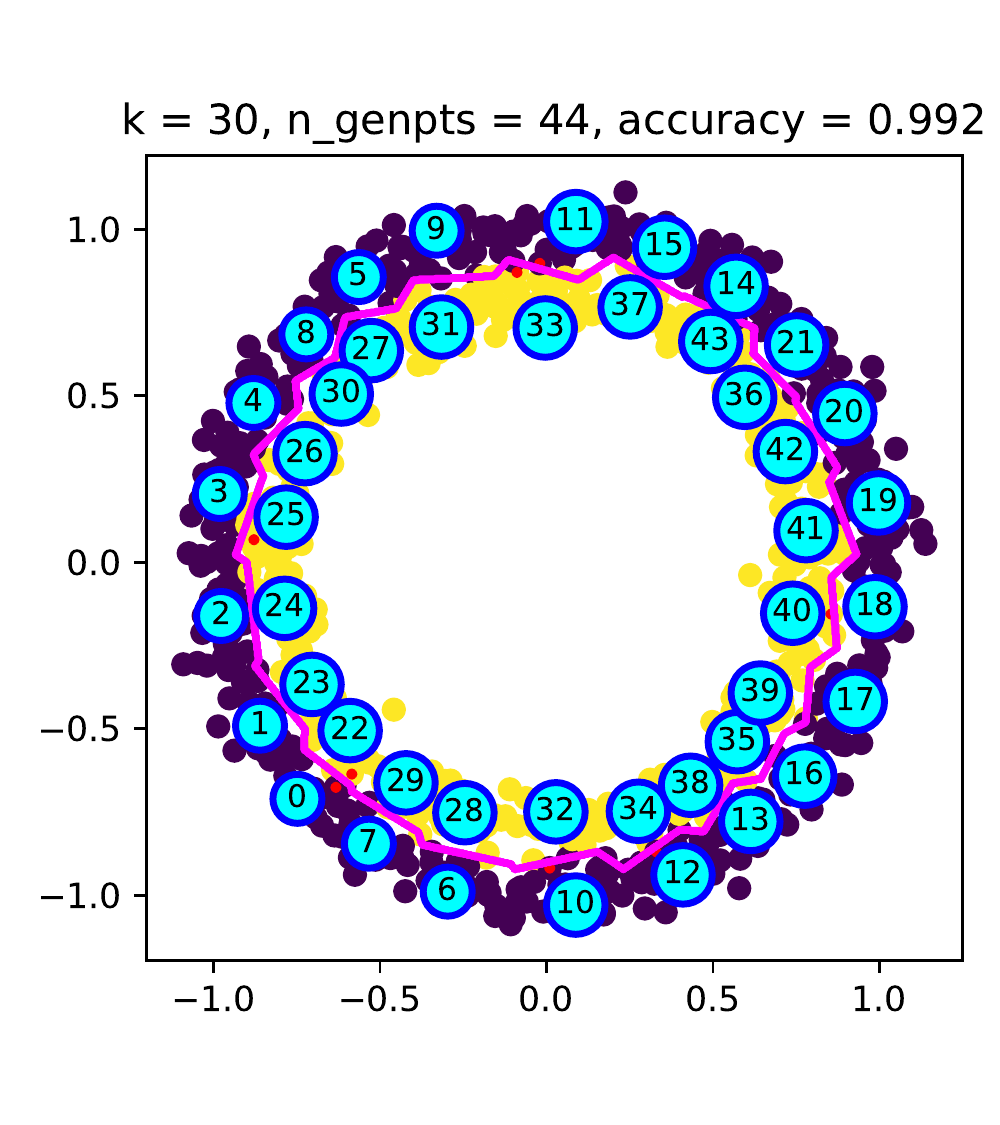}
        \caption{$k=30$}
        \label{fig:test_circles:3}
    \end{subfigure}
    \caption{Test results on synthetic circles dataset}
    \label{fig:test_circles}
\end{figure*}

\begin{figure*}
    \centering
    \begin{subfigure}[b]{0.32\textwidth}
        \centering
        \includegraphics[width=\textwidth]{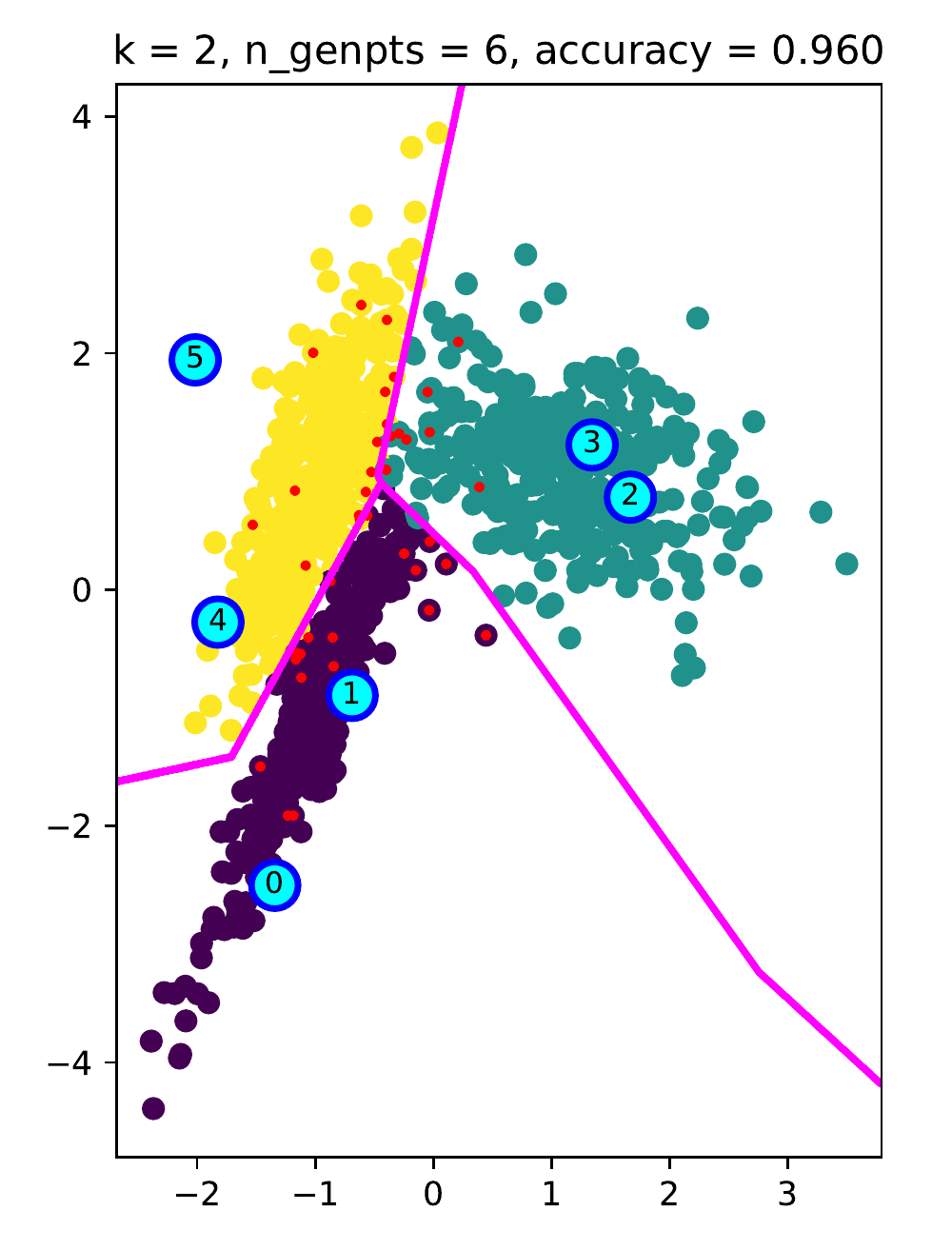}
        \caption{$k=2$}
        \label{fig:test_random:1}
    \end{subfigure}
    \begin{subfigure}[b]{0.32\textwidth}
        \centering
        \includegraphics[width=\textwidth]{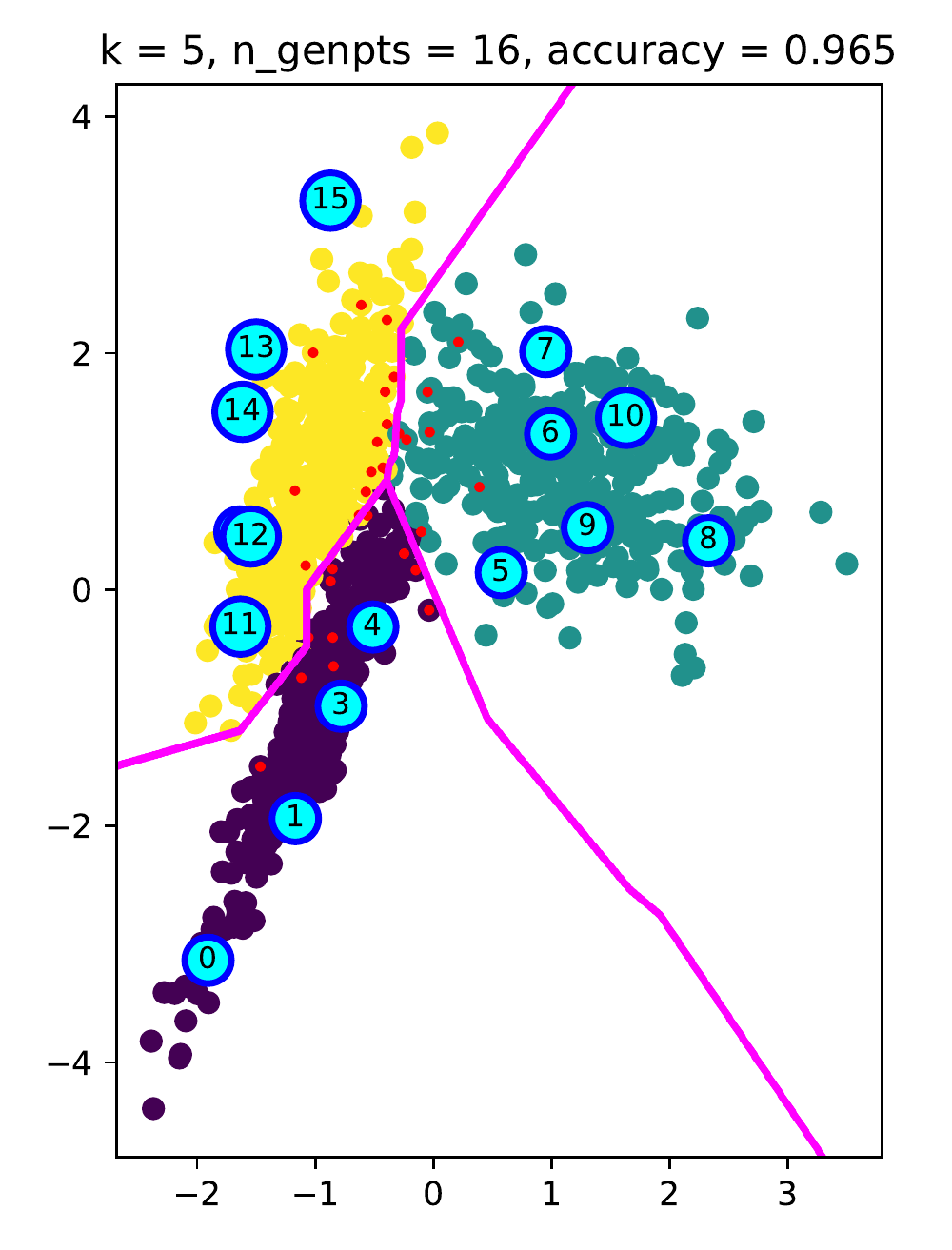}
        \caption{$k=5$}
        \label{fig:test_random:2}
    \end{subfigure}
    \begin{subfigure}[b]{0.32\textwidth}
        \centering
        \includegraphics[width=\textwidth]{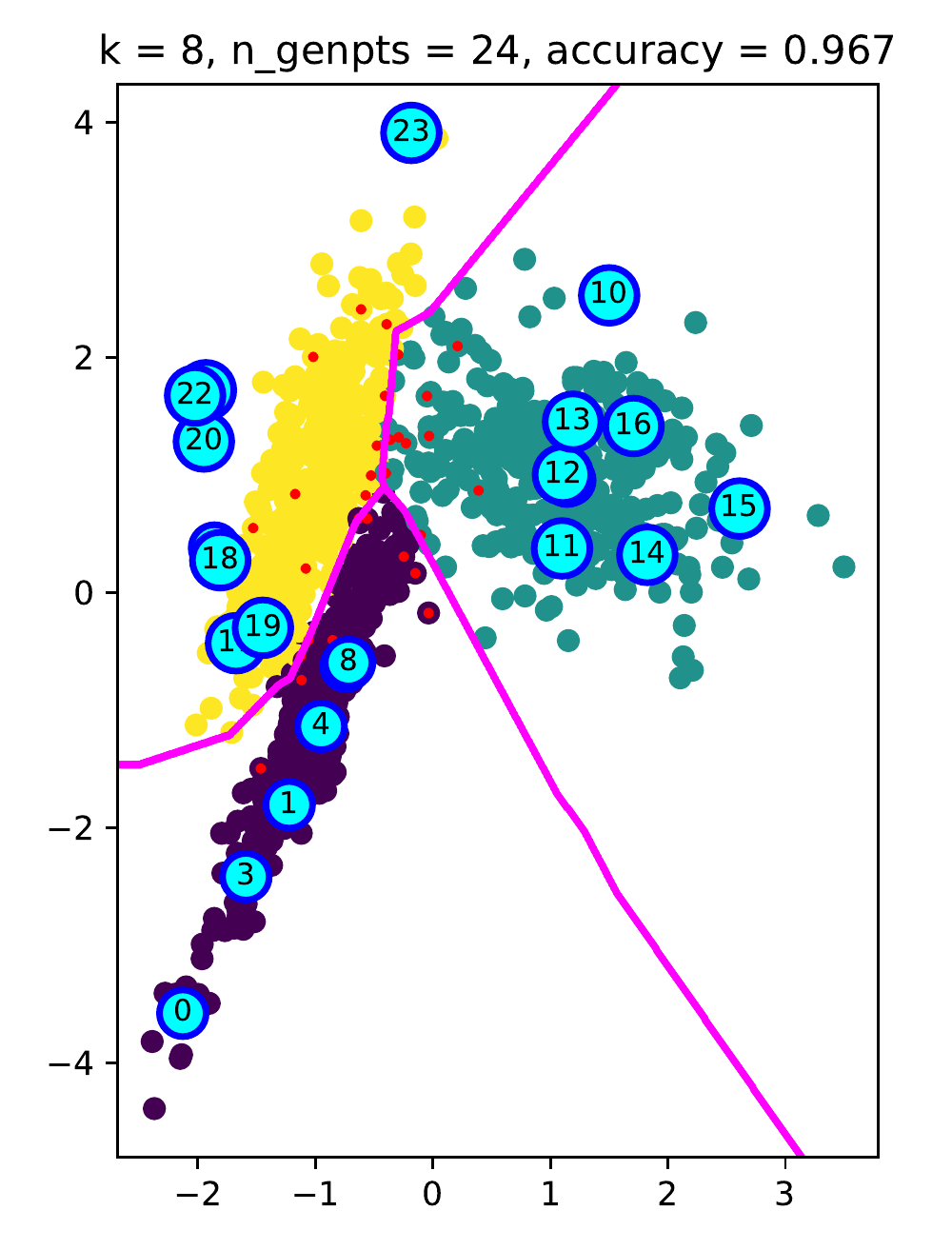}
        \caption{$k=8$}
        \label{fig:test_random:3}
    \end{subfigure}
    \caption{Tests results on randomly generated gaussians}
    \label{fig:test_random}
\end{figure*}

The Super-k algorithm is tested on synthetic datasets especially for
visualization purposes. Using the library tools, 2 class moons (Figure
\ref{fig:test_moons}), 2 class circles (Figure \ref{fig:test_circles}) and 3
class random gaussians (Figure \ref{fig:test_random}) are generated. All
datasets are processed using different values of k. The instances of different
classes are shown in distinctive colors. The misclassified instances are marked with
red dots. The generator points are shown as cyan circles with the generator point index
numbers inside. The boundaries between the generator points of different classes
are shown in magenta.

The Super-k algorithm creates PWL boundaries between different classes of data.
It can be seen on figures that increasing the number of generator points increases
the detail of the decision boundaries. On the other hand, the increase in the detail
does not always mean significantly better classification accuracy. On the moons
dataset (Figure \ref{fig:test_moons}), the accuracies for different $k$ values
are almost the same.

When the generator points from different classes come closer, their relative
orientation becomes more important. On the circles dataset (Figure
\ref{fig:test_circles}), sawtooth-like structures with sharp corners occur on
the decision boundary. For the low number of the generator points, such shapes might
become bigger and increase the classification error. During the development of
super-k algorithm, such drawbacks and possible precautions are ignored for the
sake of simplicity of the algorithm. With increasing $k$, such artifacts
become less effective and the total error decreases.

The obtained generator points do not necessarily occur on the data. This
phenomenon can be identified clearly on (Figure \ref{fig:test_random:1}). The
generator point $5$ is placed out of the data, but in competition with the
generator point $3$, it defines the boundary between the yellow and green
colored regions.

\subsection{Tests with real world datasets}
\label{sub:comparative_tests}

In this study, the SVM family of algorithms are used for comparison.
Additionally, KNN is used as another reference method. The SVM
variants\cite{fan_liblinear_2008, chang_libsvm_2011} used for the comparisons
are linear SVM (Linear SVM), SVM with Linear kernel (SVM Linear), SVM with RBF
kernel (SVM RBF), and SVM with polynomial kernel (SVM Poly). Five real world
datasets\cite{OpenML2013} (Table \ref{tbl:experimental:datasets}) are selected
for testing. Both the training (Table \ref{tbl:experimental:training_times}) and
test (Table \ref{tbl:experimental:test_times}) runs are repeated for 10 times.
In order to reduce the OS scheduler related timing noise, every run is continued
at least 1 seconds. The provided results are the mean values of the durations and
the standard deviations in parentheses. Certain parameters of the algorithms are
determined via 5-fold cross validation (CV), for every dataset separately.
Because the CV process takes long, determined parameters are saved for later
reuse.

As it can be seen from the accuracies (Table
\ref{tbl:experimental:test_accuracies}), the Super-k algorithm produces results
that are similar to the other algorithms.

When training times (Table \ref{tbl:experimental:training_times}) are
considered, the Super-k algorithm has comparable performance with the SVM family of
algorithms.

Similarly, when we compare the inference performance of the algorithms (Table
\ref{tbl:experimental:test_times}), performance of Super-k is comparable with
linear SVM, and is better than the others.

\begin{table*}
    \centering
    \caption{Experimental results with real world datasets}
    % ~\\
    \begin{subtable}[t]{\textwidth}
        \centering
        \caption{Datasets}
        \begin{tabular}{|l|rrr|}
            \hline
                        & Sample Size & Features & Train/Test Sizes \\
            \hline
            optdigits   & 5620        & 64       & 3823/1797        \\
            USPS        & 9298        & 256      & 7291/2007        \\
            satimage    & 6430        & 36       & 5144/1286        \\
            letter      & 20000       & 16       & 16000/4000       \\
            isolet      & 7797        & 617      & 6240/1557        \\
            \hline
        \end{tabular}
        \label{tbl:experimental:datasets}        
    \end{subtable}
    ~\\
    \begin{subtable}[t]{\textwidth}
        \centering
        \caption{Test accuracies}
        \begin{tabular}{|l|rrrrr|}
            \hline
                       & optdigits & USPS  & satimage & letter & isolet \\
            \hline
            Super-k    & 0.976     & 0.943 & 0.918    & 0.950  & 0.940  \\
            Linear SVM & 0.950     & 0.915 & 0.835    & 0.701  & 0.956  \\
            SVM Linear & 0.962     & 0.935 & 0.877    & 0.850  & 0.959  \\
            SVM RBF    & 0.983     & 0.954 & 0.918    & 0.979  & 0.969  \\
            SVM Poly   & 0.975     & 0.951 & 0.896    & 0.955  & 0.969  \\
            KNN        & 0.980     & 0.950 & 0.911    & 0.961  & 0.927  \\
            \hline
        \end{tabular}
        \label{tbl:experimental:test_accuracies}        
    \end{subtable}
    ~\\
    \begin{subtable}[t]{\textwidth}
        \centering
        \caption{Training times in milliseconds, Mean{\tiny (StdDev)}}
        \begin{tabular}{|l|rrrrr|}
            \hline
                       & optdigits             & USPS                   & satimage             & letter                  & isolet                  \\
            \hline
            Super-k    & 1008.9{\small (16.1)} & 5651.8{\small (232.0)} & 285.1{\small (37.7)} & 3534.4{\small (1769.6)} & 1628.0{\small (148.3)}  \\
            Linear SVM & 96.2{\small (3.9)}    & 1670.4{\small (78.0)}  & 126.0{\small (6.4)}  & 1580.2{\small (104.5)}  & 6061.3{\small (50.7)}   \\
            SVM Linear & 139.5{\small (11.5)}  & 1477.2{\small (93.3)}  & 361.7{\small (24.3)} & 3784.7{\small (163.3)}  & 5525.9{\small (142.1)}  \\
            SVM RBF    & 387.6{\small (18.6)}  & 2230.2{\small (137.9)} & 504.1{\small (8.2)}  & 15299.6{\small (834.5)} & 12867.2{\small (245.0)} \\
            SVM Poly   & 129.6{\small (1.2)}   & 1773.6{\small (213.2)} & 421.5{\small (11.3)} & 2649.8{\small (30.2)}   & 7824.1{\small (123.2)}  \\
            KNN        & 19.2{\small (1.4)}    & 165.7{\small (5.6)}    & 17.1{\small (0.8)}   & 52.5{\small (0.7)}      & 266.0{\small (3.1)}     \\
            \hline
        \end{tabular}
        \label{tbl:experimental:training_times}
    \end{subtable}
    ~\\
    \begin{subtable}[t]{\textwidth}
        \centering
        \caption{Inference times in milliseconds, Mean{\tiny (StdDev)}}
        \begin{tabular}{|l|rrrrr|}
            \hline
                       & optdigits            & USPS                   & satimage            & letter                 & isolet                  \\
            \hline
            Super-k    & 3.4{\small (0.2)}    & 12.1{\small (0.8)}     & 0.8{\small (0.1)}   & 34.1{\small (13.7)}    & 2.4{\small (0.3)}       \\
            Linear SVM & 1.8{\small (0.2)}    & 2.1{\small (0.2)}      & 1.1{\small (0.0)}   & 3.8{\small (0.3)}      & 2.8{\small (0.3)}       \\
            SVM Linear & 67.9{\small (3.1)}   & 644.2{\small (35.2)}   & 81.3{\small (2.2)}  & 1138.9{\small (37.9)}  & 4177.0{\small (219.8)}  \\
            SVM RBF    & 183.1{\small (3.5)}  & 903.4{\small (80.6)}   & 135.4{\small (4.6)} & 2523.8{\small (188.6)} & 5312.9{\small (218.9)}  \\
            SVM Poly   & 102.6{\small (12.4)} & 946.0{\small (75.0)}   & 78.5{\small (2.5)}  & 793.7{\small (65.4)}   & 4137.8{\small (253.8)}  \\
            KNN        & 290.5{\small (61.3)} & 1380.8{\small (388.0)} & 146.4{\small (3.9)} & 338.9{\small (22.6)}   & 3443.5{\small (1559.9)} \\
            \hline
        \end{tabular}
        \label{tbl:experimental:test_times}
    \end{subtable}
    % ~\\
    \label{tbl:experimental}
\end{table*}

\section{Conclusion}
\label{sec:conclusion}

A new foundational PWL classification algorithm, \textbf{Super-k}, which is inspired
from well known ideas, is introduced in this paper. The contributions of the
proposed algorithm are as follows:

\begin{itemize}
    \item A method for voxelization of multidimensional data is proposed.
    \item A simple and efficient likelihood function is introduced, and under
    some conditions, its usage in place of Euclidean distance is explained.
    \item A new approach for data classification, based on Voronoi
    tessellations, is presented.
\end{itemize}

Super-k might help the applications that have a limited computational resource.
Space systems have limited onboard resources and restrictions on the power
budget. For such systems, SVM is considered\cite{shang_fuzzy-rough_2013,
jallad_hardware_2014} as a solution. The Super-k algorithm might be an
alternative for such requirements.

The proposed algorithm can be improved in many ways and the ideas that created
the Super-k algorithm may open ways to many different solutions.

\section*{Acknowledgements}
\label{sec:acknowledgements}

This research was supported by the Turkish Scientific and Technological Research
Council (TUBITAK) under project no. 118E809.

We would like to thank the reviewers for their thoughtful comments and their
constructive remarks.

\appendix

\section{Derivation of Super-k Likelihood}
\label{apx:likelihood_derivation}

The Super-k likelihood function (\ref{eq:likelihood}) can be derived from the
logarithm of multivariate normal distribution (\ref{eq:multivariate_normal}).

\begin{align}
    p(x|\omega) &= \frac{1}{(2\pi)^{\frac{d}{2}} |\Sigma|^\frac{1}{2}}exp\left[{-\frac{1}{2}(x-\mu)^T\Sigma^{-1}(x-\mu)}\right]
    \label{eq:multivariate_normal}\\
    g_i(x) &= \ln p(x|\omega_i) \notag\\
           &= -\frac{1}{2}(x-\mu_i)^T\Sigma^{-1}(x-\mu_i) - \frac{d}{2}\ln (2\pi) - \frac{1}{2}\ln |\Sigma|
    \label{eq:log_likelihood}
\end{align}

Expanding (\ref{eq:log_likelihood}) and making $\Sigma = I$ gives

\begin{align}
    g_i(x) &= -\frac{1}{2}\left[x^T x - x^T \mu_i - \mu_i^T x + \mu_i^T \mu_i \right] - \frac{d}{2}\ln (2\pi)
    \label{eq:expanded_log_likelihood}
\end{align}

For a two-side comparison between $g_i(x)$ and $g_j(x)$, the constant terms have no
effect. Removing the terms in (\ref{eq:expanded_log_likelihood}), the constant with
respect to $i$ gives

\begin{align}
    g_i(x) &= -\frac{1}{2}\left[ - x^T \mu_i - \mu_i^T x + \mu_i^T \mu_i \right] \notag\\
    g_i(x) &= x^T \mu_i - \frac{1}{2} \mu_i^T \mu_i
    \label{eq:simplified_log_likelihood}
\end{align}

As a result of such simplifications, the Super-k likelihood
(\ref{eq:simplified_log_likelihood}) has been obtained.

When one-to-many comparisons are considered, it can be shown that maximizing the
super-k likelihood is the same as minimizing the Euclidean distance.

\begin{equation}
    \argmax_{i}{\left( \vect{x}^T \vect{p_i} - \frac{1}{2} \vect{p_i}^T \vect{p_i} \right)} 
    = \argmin_{i}{\left( \norm{\vect{x} - \vect{p_i}} \right)}
    \label{eq:superk_euclidean_equivalence}
\end{equation}

When the right hand side of (\ref{eq:superk_euclidean_equivalence}) is expanded,
\begin{align}
    &\argmin_{i}{\left( \norm{\vect{x} - \vect{p_i}} \right)} \notag\\
    &= \argmin_{i}{\sqrt{(\vect{x} - \vect{p_i})^T (\vect{x} - \vect{p_i})}} \notag\\
    &= \argmin_{i}{\sqrt{\vect{x}^T \vect{x} - \vect{x}^T \vect{p_i} - \vect{p_i}^T \vect{x} + \vect{p_i}^T \vect{p_i}}}
    \label{eq:expanded_euclidean_equivalence}
\end{align}

The square root operation is a monotonic function, hence, removing it does not
affect the ordering. Also $\vect{x}^T \vect{x}$ in
(\ref{eq:expanded_euclidean_equivalence}) is a constant with respect to $i$, thus it
is the same on both sides of the comparisons and can be removed.

\begin{align}
    &\argmin_{i}{\left( \norm{\vect{x} - \vect{p_i}} \right)} 
    = \argmin_{i}(- 2 \vect{x}^T \vect{p_i} + \vect{p_i}^T \vect{p_i})
    \label{eq:simpler_euclidean_equivalence}
\end{align}

Multiplying (\ref{eq:simpler_euclidean_equivalence}) with $-\frac{1}{2}$ only
changes the direction of the ordering, hence, $\argmin$ becomes $\argmax$
(\ref{eq:simple_euclidean_equivalence}). 

\begin{align}
    &\argmin_{i}{\left( \norm{\vect{x} - \vect{p_i}} \right)}
    = \argmax_{i}(\vect{x}^T \vect{p_i} - \frac{1}{2} \vect{p_i}^T \vect{p_i})
    \label{eq:simple_euclidean_equivalence}
\end{align}

\section{Derivation of $m_v$}
\label{apx:derive_mv}

Our aim is to approximate $k$, using only integer values. We define a parameter
$c$ such that $c^m = k$. Most of the time, the value of $c$ is not an integer. But
it is possible to define lower and upper integer bounds for $c$, such that $a
\le c \le b$, where $a, b$ are integer values. It is possible to calculate $c$,
such that $c = \sqrt[m]{k}$. Then, $a$ and $b$ can be found as $a = \left\lfloor
c \right\rfloor$ and $b = \left\lceil c \right\rceil$.

The exponent is a monotonic function for positive real numbers. We know that $c$
must be in somewhere between $a$ and $b$. Thus, taking the exponent of the bounds,
$a^m \le c^m \le b^m$ must be also true. Then, there must be a value of $m_v$
that satisfies the condition $a^{(m - m_v)} b^{m_v} = c^m$. If we restrict $m_v$
to be an integer, the condition can only be an approximation. Using that
approximation, we can derive an equation (\ref{eq:approximate_mv}) for the
approximate integer value of $m_v$.

\begin{align}
    a^{(m - m_v)} b^{m_v} &\approx c^m \notag\\
    a^m a^{-m_v} b^{m_v} &\approx c^m \notag\\
    \left(\frac{b}{a}\right) ^{m_v} &\approx \left(\frac{c}{a}\right)^{m} \notag\\
    m_v &\approx m \frac{\log{\left(\frac{c}{a}\right)}}{\log{\left(\frac{b}{a}\right)}}
    \label{eq:approximate_mv}
\end{align}

If we round the right hand side of (\ref{eq:approximate_mv}) to the nearest
integer, the equation for finding $m_v$ (\ref{eq:exact_mv}) becomes

\begin{equation}
    m_v =  \left\lfloor m \frac{\log{\left(\frac{c}{a}\right)}}{\log{\left(\frac{b}{a}\right)}} \right\rceil
    \label{eq:exact_mv}
\end{equation}
which is the number of variant features.

\newpage

\bibliographystyle{elsarticle-num} 
\bibliography{super-k_paper.bib}

\end{document}